\documentclass[10pt,twocolumn,letterpaper]{article}

\usepackage{cvpr}              
\definecolor{cvprblue}{rgb}{0.21,0.49,0.74}
\usepackage[pagebackref,breaklinks,colorlinks,allcolors=cvprblue]{hyperref}

\def\paperID{*****} 
\def\confName{CVIS}
\def\confYear{2025}

\title{Training-Free Robot Pose Estimation using Off-the-Shelf Foundational Models}

\author{Laurence Liang\\
McGill University\\
{\tt\small laurence.liang@mail.mcgill.ca}
}

\usepackage{float}
\usepackage{graphicx}

\begin{document}
\maketitle


\begin{abstract}
Pose estimation of a robot arm from visual inputs is a challenging task. However, with the increasing adoption of robot arms for both industrial and residential use cases, reliable joint angle estimation can offer improved safety and performance guarantees, and also be used as a verifier to further train robot policies. This paper introduces using frontier vision-language models (VLMs) as an ``off-the-shelf" tool to estimate a robot arm's joint angles from a single target image. By evaluating frontier VLMs on both synthetic and real-world image-data pairs, this paper establishes a performance baseline attained by current FLMs. In addition, this paper presents empirical results suggesting that test time scaling or parameter scaling alone does not lead to improved joint angle predictions. 
\end{abstract}

\section{Introduction}

Estimating a robot arm's pose from an image is a challenging task, due to a robot arm's high degrees of freedom. In particular, pose estimation only becomes practical when a very high degree of precision can be achieved. Robot arm control typically involves calculating the robot arm's end effector position through forward kinematics, and identifying subsequent poses through inverse kinematics in order to achieve a goal. The slightest offsets or errors in the joint angle estimates can propagate and render such forward or inverse kinematics tasks impractical. 

Vision-based pose estimation of robot arms is important with the rise in use cases for industrial and residential robots. Reliable pose estimation from visual inputs can complement the reliability of the robot arm's onboard sensors. In addition, vision-based pose estimation is relevant for monitoring robot arm performance and safety through video footage, and annotating robot arm poses for model training.

Existing methods for joint estimation often involve adding physical markers or pre-training models on existing data. While these approaches are highly effective, the former category of methods cannot be easily used ``in the wild", while the latter group requires model training which may not be accessible for edge model deployments.

This paper proposes the idea of using frontier vision-language foundation models (VLMs) as an ``off-the-shelf" alternative to these existing pose estimation methods. Such an ``off-the-shelf" method would be easily accessible through models made available through cloud providers and could be used immediately with no training required. Thus, the main contributions of this paper are:

\begin{itemize}
    \item Evaluating frontier VLMs on three different data categories for robot arm pose estimation, on simulation and real-world data
    \item Identifying that current VLMs can understand the approximate ``pose" of a robot arm, but are so far unreliable for precise joint angle estimates
    \item Collecting preliminary empirical results that suggest that parameter size scaling and test time scaling provide negligible improvements for reducing the joint estimation error.
\end{itemize}

\subsection{Pose Estimation}

\begin{figure*}[h]
    \centering
    \includegraphics[width=0.9\linewidth]{img/workflow.png}
    \caption{The pose estimation workflow consists of a reference prompt and a target prompt. The reference prompt contains the ground truth joint angles associated to a reference photo. The target prompt contains only a target photo of the robot arm. The vision-language model estimates the state of the target model (joint angles) and provides upper bound and lower bound estimates. }
    \label{fig:workflow}
\end{figure*}

\section{Related Work}


Early methods for vision-based pose estimation would involve adding visual (fiducial) markers on the robot arm. Such examples include AprilTag markers \cite{olson2011aprilTag} and motion capture markers (i.e. Vicon). While fiducial markers enables high precision for joint tracking, pose estimation of a robot ``in the wild" with no visual markers would not be feasible. As a result, alternative pose estimation methods would require keypoint tracking identify individual joints and to then reconstruct the angles and distances between the robot arm's joints. Keypoint estimation methods such as SIFT, FAST and ORB would be relevant for a marker-less pose estimation method \cite{Lowe2004DistinctiveIF, Rosten2008FasterAB, Rublee2011ORBAE}. 

\citeauthor{lee2020icra:dream} introduced the DREAM dataset which features both simulation and real-world image and joint angle pairs of different robot arms. DREAM has subsequently been used as a standard benchmark for robot pose estimation. Accordingly, \citeauthor{lee2020icra:dream} used the ``perspective-n-point" to obtain joint angle estimates. 

\citeauthor{labbe2021singleviewrobotposejoint} introduced RoboPose as a ``render and compare" approach to pose estimation. During model training, RoboPose iteratively uses a renderer and a refiner to estimate the joint angle of cropped components from a robot arm image. \citeauthor{goswami2025robopeppvisionbasedrobotpose} introduced RoboPEPP which achieved new state of the art performance on the DREAM dataset. RoboPEPP first pretrains model embeddings using image masking, and then estimates the joint angles using the ``perspective-n-point" (PnP) algorithm. 

Other relevant work includes PoseLess which achieves generalizable image-to-joint prediction using synthetic data \cite{dao2025poselessdepthfreevisiontojointcontrol}, and Active Pose that uses CAD views to improve a vision-language model's (VLM) estimate of a robot arm's pose \cite{liu2025activeposeactive6dobject}.







\section{Method}

The pose estimation method uses a ``reference-target" approach, a one-shot prompting strategy with frontier VLMs to estimate the joint angles of the Franka Emika Panda robot arm. 

\subsection{Dataset}

\begin{figure}[htb]
    \centering
    \begin{subfigure}{0.48\columnwidth} 
        \includegraphics[width=\textwidth]{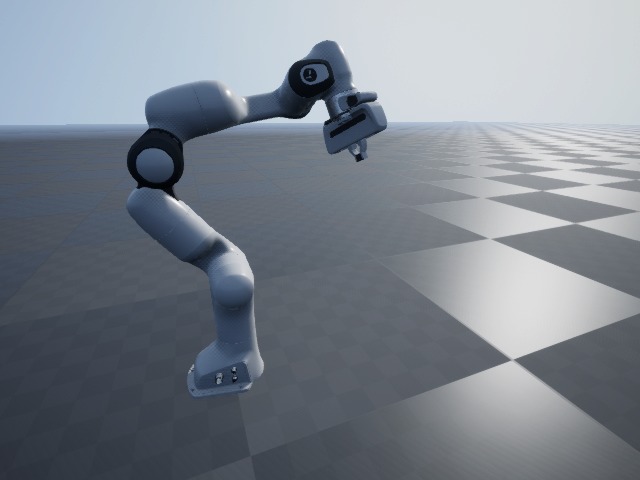}
        \caption{Simulation Subset}
        \label{fig:subfigA}
    \end{subfigure}
    \hfill 
    \begin{subfigure}{0.48\columnwidth} 
        \includegraphics[width=\textwidth]{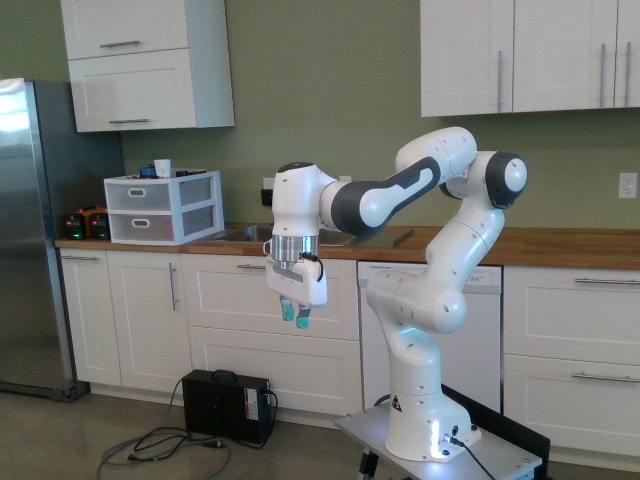}
        \caption{Realsense Subset}
        \label{fig:subfigB}
    \end{subfigure}
    \begin{subfigure}{0.48\columnwidth} 
        \includegraphics[width=\textwidth]{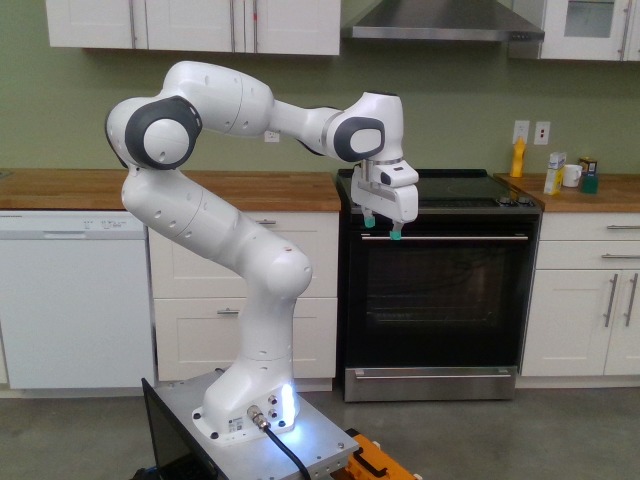}
        \caption{Orb Subset}
        \label{fig:subfigB}
    \end{subfigure}
    \caption{The three subsets of the DREAM dataset used for pose estimation \cite{lee2020icra:dream} These subsets feature the Franka Emika Panda arm. }
    \label{fig:snapshot}
\end{figure}

This paper assembled a new dataset, DREAM-Mini, which is a subset of the DREAM dataset \cite{lee2020icra:dream}. The DREAM dataset contains views of robot arms in simulation and in the real world, with ground truth measurements for each joint. DREAM-Mini maintains three subsets of the original DREAM dataset that involve the Franka Emika Panda robot arm: (1) the simulation subset (S1), (2) the Realsense subset (S2) which contains a single camera view in the real world, and (3) the Orb subset (S3) that contains three camera views in the real world. (Fig. \ref{fig:snapshot}) The simulation subset contains 63 images, the Realsense subset contains 49 images, and the Orb subset contains 49 images. Each subset also has a corresponding reference image that is taken from the same parent set.

\begin{table*}[h]
\centering
\caption{Frontier Model Performance on the DREAM-Mini Dataset for Panda Arm Pose Estimation (Units in Radians Unless Specified Otherwise)}
\resizebox{\linewidth}{!}{
\begin{tabular}{l|ll|ll|lll}
\toprule
subset               & panda\_sim\_full\_view &              & panda\_realsense\_full\_view &              & panda\_orb\_full\_view &              &                              \\
\midrule
model                & gpt-5.1                & gemini-3-pro & gpt-5.1                      & gemini-3-pro & gpt-5.1                & gemini-3-pro & sonnet-4.5 (1,000 R tokens)  \\
\midrule
average error        & 0.657                  & 0.652        & 0.48                         & 0.504        & 0.468                  & 0.578        & 0.416                        \\
average uncertainty  & 0.286                  & 0.189        & 0.224                        & 0.19         & 0.281                  & 0.227        & 0.086                        \\
average within bound & 41.07\%                & 30.44\%      & 39.94\%                      & 37.20\%      & 39.36\%                & 32.38\%      & 16.72\%                      \\
\midrule
joint 1              & 0.722                  & 1.165        & 0.668                        & 0.643        & 0.424                  & 1.131        & 0.678                        \\
joint 2              & 1.320                  & 0.885        & 0.242                        & 0.368        & 0.426                  & 0.427        & 0.282                        \\
joint 3              & 0.519                  & 0.426        & 0.614                        & 0.612        & 0.415                  & 0.585        & 0.361                        \\
joint 4              & 0.446                  & 0.374        & 0.375                        & 0.359        & 0.409                  & 0.440        & 0.276                        \\
joint 5              & 0.733                  & 0.618        & 0.435                        & 0.435        & 0.566                  & 0.559        & 0.490                        \\
joint 6              & 0.333                  & 0.467        & 0.423                        & 0.458        & 0.522                  & 0.579        & 0.384                        \\
joint 7              & 1.187                  & 1.280        & 0.600                        & 0.648        & 0.513                  & 0.328        & 0.441                        \\
joint 8              & 0                      & 0            & N/A                          & N/A          & N/A                    & N/A          & N/A     \\              \bottomrule      

\end{tabular}
}
\end{table*}

\begin{figure}[h]
    \centering
    \includegraphics[width=0.7\linewidth]{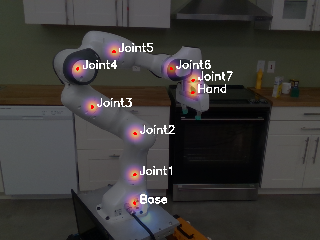}
    \caption{Labels for each joint of the Franka Emika Panda arm from the DREAM dataset. Image credit to \citeauthor{lee2020icra:dream}}
    \label{fig:labels}
\end{figure}

As shown in Fig. \ref{fig:labels}, the Franka Emika Panda robot contains 7 labeled joints. Each joint angle is measured in radians. To prevent an arbitrary reference coordinate system, each subset of DREAM-Mini has a reference image with ground truth joint angles. As a result, given these sets of ground truth angles, any new target image's angles become non-arbitrary and unique. 

To leverage the uniqueness of joint angles using a ``reference-target" setup, this paper proposes a one-shot prompting strategy that includes the ground truth reference pose, and aims to predict the target image's pose. (Fig. \ref{fig:workflow}) The full prompts are included in the Appendix section. 

The VLM is prompted to predict each joint's angle in radians, and is also asked to provide a lower bound and upper bound estimate. This ``bounded estimate" serves to measure a VLM's ability to narrow the preciseness of its estimate, while also indicating whether the model is able to understand the uncertainty associated with its estimate. 

The main metric is the mean angle error in radians, at each joint between the ground truth and the estimate. This can be formalized as:

\begin{align}
    e &= \frac{1}{M}\frac{1}{N}\sum_{i=1}^M\sum_{j=1}^N|\hat{q}_{i,j}- q_{i,j}|
\end{align}

where $\hat{q}_{i,j}$ is the estimated i-th joint angle in radians of the robot arm for the j-th sample, while $q_{i,j}$ is the corresponding ground truth. $e$ is the mean error. Another metric, ``average within bound", is the percentage of ground truth estimates that overlap with the ground truth joint angle. 

\subsection{Model Selection}

This paper looks at current frontier models, grouped in the following categories. All the models were sourced from OpenRouter. Qwen-3-VL models used Novita as a provider through OpenRouter.

\subsubsection{Frontier, Closed Source Multi-Modal Models} 

GPT-5.1, Gemini-3-Pro, and Sonnet-4.5 are evaluated as frontier, closed source VLMs \cite{gpt5-1, gemini-3, sonnet4-5}.

\subsubsection{Reasoning Models: Test Time Scaling} 

Sonnet-4.5 is used to evaluate whether increasing reasoning token count can be used to improve task performance, which is known as test-time scaling. \cite{muennighoff2025s1simpletesttimescaling}

\subsubsection{Parameter Size Scaling}

Qwen3-VL models are used to determine the effects of parameter scaling. \cite{kaplan2020scalinglawsneurallanguage} The Qwen3-VL model family have openly accessible model weights: the 30B and 235B parameter ``thinking" models are used. \cite{Qwen-VL, bai2025qwen3vltechnicalreport, yang2025qwen3technicalreport}

\section{Results}

\subsection{Frontier Model Performance}

Table 1 shows frontier closed source model performance on the three DREAM-Mini subsets. The average error is quite large across all three subsets and across different frontier VLMs. While average errors from 0.4 radians (23 degrees) to 0.65 radians (37 degrees) is sufficiently reliable to approximate the pose of the robot, the current models are too inaccurate to be used as an alternative for forward or inverse kinematics. 

The joint-specific error distribution also varies between subtasks. This could likely be explained by highly different camera field of views between the subtasks, suggesting that the camera pose and constraints on the robot's overall pose can affect joint-specific error estimates. 

The average error across subtasks are comparably large, which suggests that VLMs do not have differing performances between simulation or real-world data. Additionally, less than 42\% of joint estimates fall within the error bound, thus suggesting that VLMs have trouble understanding ``uncertainty" in their estimates. This is evident for Sonnet-4.5, which provides a comparatively smaller error bound estimate, but has a much lower success rate at providing error bounds that overlap with the ground truth joint angle. 

\subsection{Test Time Scaling}

\begin{table}[h]
\centering
\caption{Scaling the Reasoning Token Count for Sonnet-4.5 on the Orb View Subset}
\resizebox{\columnwidth}{!}{
\begin{tabular}{l|l|l|l}
\toprule
sonnet-4.5           & 1,000 tokens           & 5,000 tokens & 10,000 tokens  \\
\midrule
average error        & 0.416                  & 0.392        & 0.405          \\
average uncertainty  & 0.086                  & 0.09         & 0.106          \\
average within bound & 16.72\%                & 17.49\%      & 17.49\%        \\
\midrule
joint 1              & 0.678                  & 0.619        & 0.656          \\
joint 2              & 0.282                  & 0.279        & 0.270          \\
joint 3              & 0.361                  & 0.346        & 0.385          \\
joint 4              & 0.276                  & 0.291        & 0.275          \\
joint 5              & 0.490                  & 0.446        & 0.448          \\
joint 6              & 0.384                  & 0.394        & 0.439          \\
joint 7              & 0.441                  & 0.367        & 0.364          \\
joint 8              & N/A                    & N/A          & N/A         \\
\bottomrule
\end{tabular}
}
\end{table}

Table 2 shows Sonnet-4.5 performance subject to different reasoning token budgets. These preliminary results suggest that increasing the reasonging token budget does not lead to any significant improvement in pose estimation. 

\subsection{Parameter Size Scaling}

\begin{table}[h]
\centering
\caption{Parameter Size Scaling for Qwen-3-VL Thinking on the Orb View Subset}
\begin{tabular}{l|l|l}
\toprule
qwen-3-vl            & 30B    & 235B     \\
\midrule
average error        & 0.395  & 0.553    \\
average uncertainty  & 0.047  & 0.145    \\
average within bound & 5.80\% & 16.67\%  \\
\midrule
joint 1              & 0.246  & 0.728    \\
joint 2              & 0.389  & 0.353    \\
joint 3              & 0.363  & 0.57     \\
joint 4              & 0.550  & 0.493    \\
joint 5              & 0.444  & 0.626    \\
joint 6              & 0.490  & 0.638    \\
joint 7              & 0.285  & 0.462    \\
joint 8              & N/A    & N/A     \\
\bottomrule
\end{tabular}
\end{table}

Table 3 evaluates how increasing model size affect pose estimation performance. Interestingly, for the Qwen3-VL thinking series, a larger model has a higher average error. However, given that the smaller 30B model already has a large error, the increase in model size does not imply that there exists an inverse correlation between model size and state estimation performance. Rather, these results suggest that scaling parameter size alone is not sufficient to attain reliable pose estimates. 

\section{Conclusion \& Discussion}

\subsection{Main Findings}

The objective of this paper was to determine whether frontier VLMs can be used ``off-the-shelf" to estimate the joint angles of a robot arm from a single image. By curating the DREAM-Mini dataset, this paper evaluated frontier VLMs on simulation and real-world data. The results suggest that current frontier VLMs can estimate the ``approximate" pose of a robot arm, but struggle to achieve highly precise estimates. Furthermore, the results suggest that test time scaling and parameter size scaling are insufficient to achieve better estimates. Furthermore, the error bound estimates suggest that frontier VLMs struggle to understand smaller pose differences and the uncertainty involved in their estimates. 

\subsection{Limitations}

The curated DREAM-Mini dataset is small and only uses a single robot arm. While the results are sufficient to establish a baseline, a larger dataset with more diverse robot arm types would be desirable. 

A further study of different prompting methods, such as self-refinement, few-shot prompting and tool-calling would be relevant to further determine frontier VLM capabilities for reliable pose estimation.

\subsection{Applications}

One relevant application is using robot arm pose estimation as a spatial reasoning benchmark for frontier models. If VLMs can achieve more precise pose estimates, future applications can include serving as a verifier for reward modeling (in reinforcement learning), monitoring robot arms performance and safety through video feeds, and serving as a redundant pose estimation system to complement the robustness of existing sensor-based methods.

If models attain a high degree of accuracy for pose estimation, this would imply that models can estimate many states of a robot from a single image. It would be crucial to evaluate model capabilities in the context of model alignment and safety, to ensure that VLMs do not infer sensitive information from robot arm photos that should be out of reach. 

\subsection{Future Directions}

Future work should investigate additional training-free methods such as few-shot prompting, self-refinement, and tool calling. It would also be relevant to explore post-training methods, such as reinforcement learning fine tuning (RFT), to determine if models can be ``steered" to be better aligned with their pose estimation capabilities. 

{
    \small
    \bibliographystyle{ieeenat_fullname}
    \bibliography{main}
}


\appendix

\section{Reproducibility Statement}

The code and data access instructions are available at the following  repository: \url{https://anonymous.4open.science/r/robot-pose-estimation-vlm-AB8B/README.md}.




\section{Prompt}

The following are used as the reference prompt and as the target prompt.

\subsection{Reference Prompt}

\begin{quote}
``You are a robot arm pose estimator. The following reference image has these joint angles in radians:

\{joint\_angles\_str\}

The user will provide you with another image and you must estimate the joint angles in radians, the upper bound estimate and the lower bound estimate as a string of format 'angle/lower\_bound\_angle/upper\_bound\_angle'. Return ONLY a JSON object with the same structure (joint names as keys, positions in radians as values)."

\end{quote}

\subsection{Target Prompt}

\begin{quote}
Estimate the joint angles for this robot arm image:
\end{quote}

\end{document}